\documentclass{article}
\usepackage{spconf,amsmath}
\usepackage{cite}
\usepackage{amssymb,amsfonts}
\usepackage{algorithmic}
\usepackage{textcomp}
\usepackage{xcolor}
\usepackage{balance}
\usepackage{color}
\usepackage{multirow}
\usepackage{url}
\usepackage{booktabs}
\usepackage{array}
\usepackage{caption}
\usepackage{graphicx, subfig}

\newcommand\etal{\emph{et~al.}}

\title{A Real-Time Deep Network for Crowd Counting}
\name{Xiaowen Shi$^{\dagger}$, Xin Li$^{\dagger\ddagger}$, Caili Wu$^{\dagger\ddagger}$, Shuchen Kong$^{\star}$, Jing Yang$^{\dagger}$, Liang He$^{\dagger\ddagger}$}
\address{$^{\ddagger}$Shanghai Key Laboratory of Multidimensional Information Processing\\
$^{\dagger}$East China Normal University, Shanghai, China\\
$^{\star}$Videt Tech Ltd., Shanghai, China}

\begin{document}
\maketitle

\begin{abstract}
Automatic analysis of highly crowded people has attracted extensive attention from computer vision research. Previous approaches for crowd counting have already achieved promising performance across various benchmarks. However, to deal with the real situation, we hope the model run as fast as possible while keeping accuracy. In this paper, we propose a compact convolutional neural network for crowd counting which learns a more efficient model with a small number of parameters. With three parallel filters executing the convolutional operation on the input image simultaneously at the front of the network, our model could achieve nearly real-time speed and save more computing resources. Experiments on two benchmarks show that our proposed method not only takes a balance between performance and efficiency which is more suitable for actual scenes but also is superior to existing light-weight models in speed.
\end{abstract}


\section{Introduction}
\label{sec:intro}

By analyzing and understanding the crowd behavior and congestion levels in detail, some preventable calamities such as the stampede could be alleviated, which make great sense for public security. A strong demand to develop a responsive and efficient crowd counting application to effectively control the harm of emergencies is increasing and brings a big challenge to this vision task.

The existing methods to address crowd counting problem could be divided into two groups: count-oriented approaches and density-oriented approaches. Count-oriented approaches simply output the number of people by using a detector to detect objects in a sliding window that glides across the entire image. However, when the density of crowd is extremely dense, the spatial distributions are almost totally different in each image, which makes count-oriented approach invalid. In this way, spatial information is displayed {color{red} in the form of}through the density map to indicate the amount of people across the whole image. This density map provides more accurate and comprehensive information, which could be a crucial part of making correct decisions in highly varied crowded scenes.

With recent development of the convolutional neural network (CNN), researchers employ CNN to accurately estimate the crowd count from images or videos \cite{sam2017switching,sindagi2017generating,walach2016learning,wu2018adaptive,wu2019video}. However, it is always challenging to deal with scale variations on static images, especially in diversified scenes such as different camera perspectives and irregular crowd clusters. For this reason, many previous works investigate the multi-scale architectures~\cite{boominathan2016crowdnet,sam2017switching,sindagi2017generating,zhang2016single } as the backbone to deal with this problem. Although they outperform than other kinds of method yet, this kind of complex models with a large number of parameters will probably cause time-consuming and sub-optimal problem, which would be inappropriate for applications need fast response. To sum up, it is still far from the balance of accuracy and efficiency required in actual scene.

In this paper, we propose the compact convolutional neural network (C-CNN) to simplify multi-branch of CNN model. It involves a small number of parameters and achieves satisfying performance. The network utilizes three filters with different sizes of local receptive field in one layer. The generative feature maps are merged directly after receptive fields, and then fed into a CNN structure to fit a density map.Compared with existing methods, the proposed model achieves substantially enhanced performance with faster speed and the maintain of accuracy.

The contributions of this work are as follows:

\begin{itemize}
\item Our network is simple enough to be trained efficiently when compare with those multi CNN frameworks 
    which need pre-trained model in each branch. In addition, 
    it requires less computing resources and is more practical.
\item We obtain an optimal balance between the efficiency of the model and the accuracy of the estimated count, ensuring that our model can achieve the accurate results effectively.
\end{itemize}

\section{Related Work}
\label{sec:related}

\textbf{Traditional approaches.}
The early researches~\cite{dollar2012pedestrian} adopted a detection-style framework to carry out the function of counting. These methods detected the existence of a pedestrian in a sliding window by training a classifier using features extracted from a complete pedestrian. But it is difficult to count the exact number of people if most of the target objects are seriously obscured in highly congested scenes. In this case, researchers began to use specific body parts features to construct boosted classifiers ~\cite{wu2007detection}. Although the detection-based approaches have been improved though this modification, the perform is still poor in extremely dense situation, so researchers tried to design regression-based approaches to directly map the features extracted from image patches to scalar values~\cite{chan2009bayesian,idrees2013multi}.

Nevertheless, regression-based methods can not perceive crowd distributions as they ignored important spatial information and regressed on the global count. Density estimation-based approaches are therefore developed with the ability to conduct pixel-wise regressions. Linear mapping~\cite{lempitsky2010learning} and non-linear mapping~\cite{pham2015count} methods were utilized for density calculation successively.

\vspace{0.08in}
\noindent\textbf{CNN-based methods.}
With the breakthrough of deep learning in computer vision~\cite{krizhevsky2012imagenet}, some researchers tried to use convolutional neural network as feature extractor for crowd counting task ~\cite{zhang2016single, sam2017switching, babu2018divide}. They adopted multiple CNN branches with different receptive fields to enable multi-scale adaptation and then combined the output feature maps of different level of a congested scene and mapped them to a density map. These methods exactly obtained excellent performance on the highly congested scene, but they need to pre-train each single-network for global optimization. Also, the branch structure for learning different features for each column is inefficient, the redundant parameters have a negative impact on the final performance. Moreover, this kind of model is inactive in real-world because of the low speed and high latency in inference. As a remedy, single-branch counting networks with scale adaptations were proposed. Cao~\etal~\cite{cao2018scale} computed high-quality maps with a new encoder-decoder network, as well as a SIM local pattern consistent loss. However, it still suffers from a large number of parameters.

Unlike approaches mentioned above, the work in this paper is specifically aimed at reducing the number of parameters of the network by designing a sparse network structure. Specifically, we use three stacked filers of different size and directly target a merged feature map at once. In this way we can utilize sparsity at the filter level to optimize parallel computing and increase network adaptability to scale, making it spend less time on training and optimization.

\section{Proposed Approach}
\label{sec:method}

 \begin{figure}[t]
  \centering
  \includegraphics[width=\linewidth]{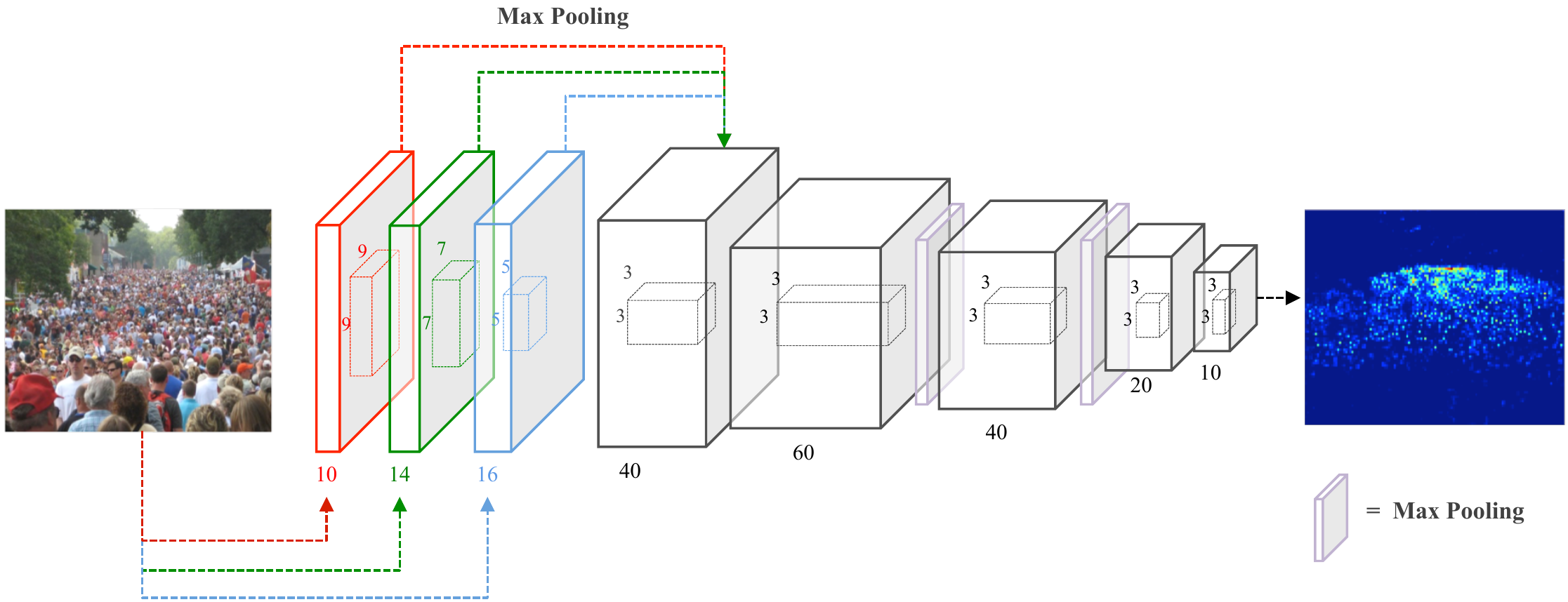}
  \caption{Overview of the proposed C-CNN architecture. The network incorporates three parallel filters of different size in the color of red, green and blue respectively, which are merged to estimate crowd density estimation.}
  \label{fig:framework}
\end{figure}

\subsection{Compact Convolutional Neural Network}

In a typical multi-scale CNN architecture, features extracted from multi-column CNN with different size of receptive fields. However, we perform multiple convolutions in parallel on the input image and combine all the output results into a very deep feature map in order to avoid parameters of our model increase explosively.

The overall architecture of our framework is illustrated in Figure~\ref{fig:framework}. The network could be divided into two components: the parallel convolution layer with different kernels and the convolution or pooling layers that followed. In the front part, the red layer in the figure \ref{fig:framework} is designed to pay more attention to large receptive fields. The green one is on behalf of dense crowds and the blue one stands for highly congested crowded scenes. All of them are followed by a 2 $\times$ 2 max-polling layer. After the extraction process of various receptive fields, the feature maps are merged as feature fusion for follow-up layers to perform down-sampling. We find that using only one layer of convolution is enough to extract different spatial features and could ameliorate the efficiency of feature extraction from multiple branches. This is why the network is faster and accurate. The latter part consists of 6 convolutional layers specifically. Note that the third layer and the fourth layer are followed by a max-pooling layer with another 2 $\times$ 2 kernel size. The last convolution layer uses 1 $\times$ 1 filter to aggregate the feature maps into a density map.

Compared with those multi-column CNNs, our method has many improvements. Firstly, we find that merging feature maps after first receptive fields in the head of the network outperforms connecting continued convolutional operations. Through our analysis, we consider that only using one layer of convolution could extract more comprehensive details of images, while the whole convolutional neural network does well on capturing local features but it would disrupt the spatial information. Another advantage is that comparing with the multi-column architecture which is always puzzled by the redundancy and repetition of the number of filters in each column, our approach can be seen as discarding the extra convolution operations. Thirdly, our approach is proved to make a reasonable trade-off between model performance and the number of network parameters. The experiments validate the effectiveness of the proposed structure.

\subsection{Implementation Details}
\label{section:implementation details}

\subsubsection{Ground truth generation}
\label{section:ground Truth}
We use the geometry-adaptive kernels to generate a ground truth density map of the highly congested scenes. Adapting a Gaussian kernel each head annotation becomes vague, so that we generate the ground truth density maps with the spatial distribution information across the whole image. This method alleviates the difficulty of the regression because we could get more accurate and comprehensive information rather than predict the exact point of head annotation directly. The geometry-adaptive kernels are defined as $$F(x) = \sum\limits_{i=1}^N \delta (x - x_i) \times G_{\sigma_i}(x),$$ where ${\sigma _i} = \beta {\bar d_i}$, $x_i$ stand for a target object in the ground truth $ \delta$, and convert ${\delta (x - {x_i})}$ into a density map, we convolve this with a Gaussian kernel with a standard deviation of $ \sigma_i$. Here $\bar d_i$ means the average distance of k nearest neighbors of target object $x_i$. In this experiment, we create density maps with $\sigma_i$ = 15.

\subsubsection{Training details}

The feature maps output from our model are mapped to the density maps adopting filters of size 1 $\times$ 1,  then we use Euclidean distance to measure the difference between the output density map and the corresponding ground truth. Here we define the loss function as follows, $$L(\Theta)=\frac{1}{N}\sum\limits_{i=1}^N ||f({X_i},\Theta)-F_i||_2,$$ Where $F_i$ is the ground truth density map of image $ X_i$, the ${f({X_i},\Theta )}$ is an estimated density map which is parameterized with $\Theta$ for the sample $ X_i$. During training, we set batch size to 8 and use Adam with learning rate of $10^{-5}$.

\section{Experiments}
\label{sec:exp}

\subsection{Results and Comparison}
We conduct a comprehensive study using the ShanghaiTech dataset~\cite{zhang2016single}and The WorldExpo'10 dataset~\cite{zhang2015cross}. We denote our approach as C-CNN in the following comparisons and use the MAE and MSE as evaluation metric.

\begin{figure}[t]
  \centering
  \includegraphics[width=\linewidth]{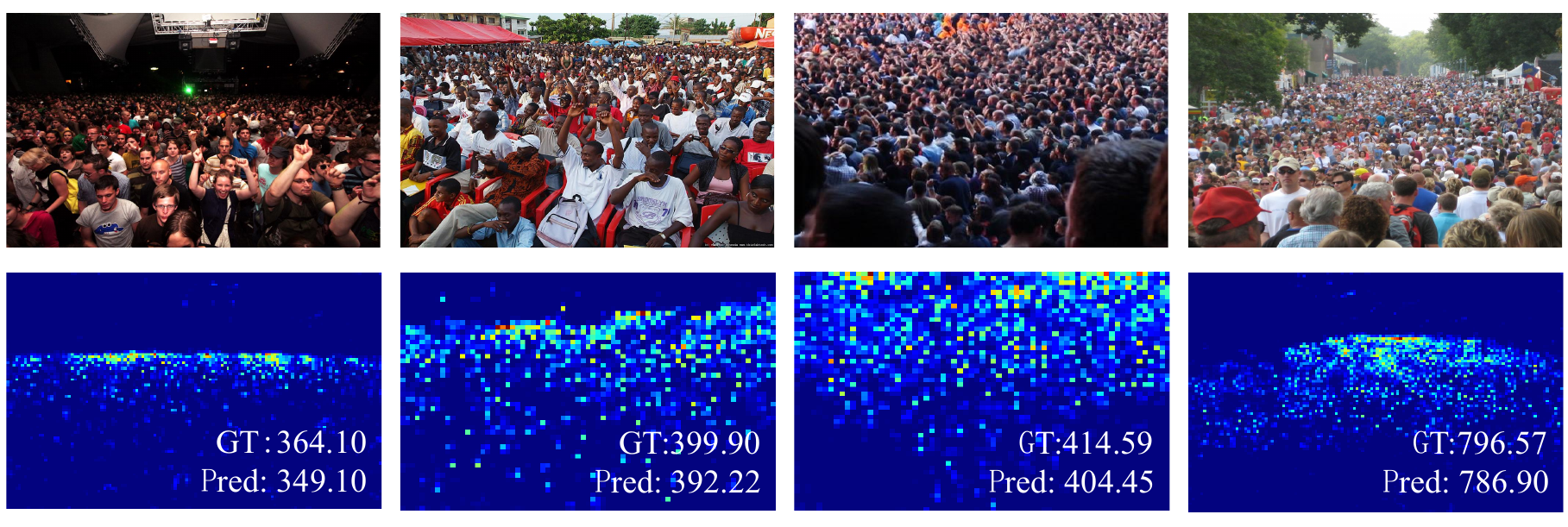}\\
  (a) ShanghaiTech Part A\\
  \vspace{0.05in}
  \includegraphics[width=\linewidth]{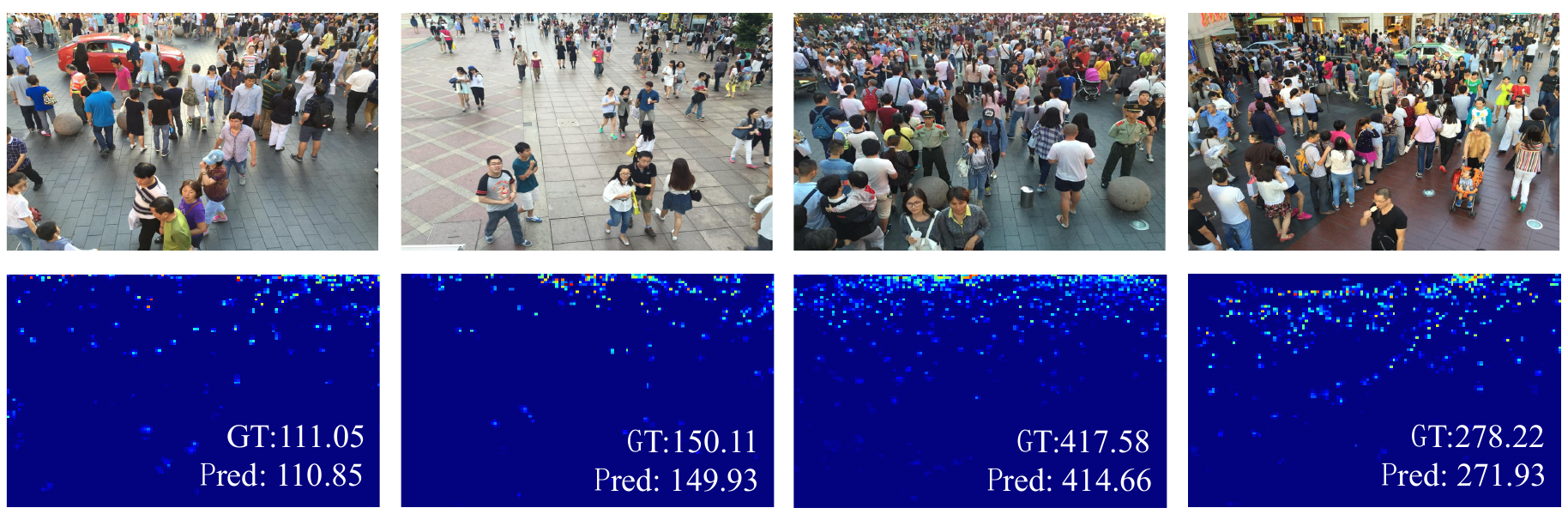}\\
  (b) ShanghaiTech Part B\\
  \caption{Qualitative results on the benchmarks.}
  \label{fig: Qualitative}
\end{figure}

\vspace{0.08in}
\noindent\textbf{ShanghaiTech dataset}~\cite{zhang2016single} is with 1198 images and 330,165 annotated people. The dataset consists of Part A and Part B. Part A includes 482 crowd images with 300 training images and 182 testing images, while Part B contains 716 images which divided into a training set with 400 images and testing set with 316 images. First, we evaluate and compare our method with other four lightweight networks and the results are shown in the upper part of Table~\ref{table:shanghai}. It displays that C-CNN with simple architecture achieves the lowest MAE in Part A and both of the lowest MAE and MSE in Part B. Note that the parameter size of our model is still the smallest one. We also compare with some large network and the results in the bottom of Table~\ref{table:shanghai}. Although the deeper models achieve better performance, their parameter size is around 200 times more than ours. Some qualitative results are presented in Figure~\ref{fig: Qualitative}.

\begin{table}
\centering
\caption{Comparison on ShanghaiTech dataset.}
\label{table:shanghai}
\small{
\begin{tabular}{lcccccc}
\hline
\multicolumn{1}{l}{\multirow{2}{*}{Method}} & \multicolumn{2}{c}{Part A} & \multicolumn{2}{c}{Part B}& Parameter  \\ \cline{2-5}
\multicolumn{1}{c}{} & MAE & MSE         & MAE          & MSE   & size      \\  \cline{1-6}
CMTL~\cite{sindagi2017cnn} & 101.3 &152.4 &20.0& 31.1&2.36M\\
Zhang ~\etal ~\cite{zhang2015cross} & 181.8        & 277.7        & 32.0         & 49.8  &0.62M \\
MCNN~\cite{zhang2016single}                   & 110.2         & 173.2          & 26.4        & 41.3 & 0.15M  \\
TDF-CNN~\cite{sam2018top} & 97.5 & 145.1          & 20.7         & 32.8   &0.13M \\
C-CNN  & \textbf{88.1}         &  \textbf{141.7}      &   \textbf{14.9}         &  \textbf{22.1}   & \textbf{0.07M} \\ \hline
ACSCP~\cite{shen2018crowd} & 75.7  &  \textbf{102.7}  &  17.2  &  27.4  &  5.10M  \\
 \footnotesize{Switching CNN~\cite{sam2017switching}} &90.4 & 135.0 & 21.6 & 33.4 & 15.30M\\
CSRNet~\cite{li2018csrnet}  &  \textbf{68.3}  &  115.0  &  \textbf{10.6}  & \textbf{16.0}  & 16.26M \\
SaCNN~\cite{zhang2018crowd}  &  86.8  &  139.2  &  16.2  &  25.8  &  24.06M \\
CP-CNN~\cite{sindagi2017generating} & 73.6 & 106.4 & 20.1 & 30.1 & 68.40M \\
\hline
\end{tabular}
}
\end{table}

\vspace{0.08in}
\noindent\textbf{WorldExpo'10 dataset} \cite{zhang2015cross} contains 1,132 annotated images that were captured in the 2010 Shanghai WorldExpo by 108 surveillance cameras. The dataset is divided into 5 different scenes, marked as S1 to S5 in Table ~\ref{table:we}. Our C-CNN delivers the best results in scene 3 and scene 4 when comparing it with other light models, displayed in the top part of the table. Besides, it also achieves the best accuracy on average with the smallest parameter size. Compare with large networks, our approach outperforms them in scene 4 with even tens of times smaller parameters. This representation reinforces the fact that C-CNN is light enough without sacrificing too much accuracy.

\begin{table}
\centering
\caption{Comparison of C-CNN with both lightweight and large networks on WorldExpo'10 dataset.}
\label{table:we}
\small{
\begin{tabular}{p{0.84in}p{0.14in}<{\centering}p{0.14in}<{\centering}p{0.14in}<{\centering}p{0.14in}<{\centering}p{0.14in}<{\centering}p{0.20in}<{\centering}c}
\hline
Method & S1 & S2         & S3          & S4  & S5          & Avg. & Params     \\  \cline{1-8}
Zhang ~\etal ~\cite{zhang2015cross} & 9.8 & \textbf{14.1} & 14.3 & 22.2 & 3.7 & 12.9 & 0.62M \\
MCNN~\cite{zhang2016single} & 3.4 & 20.6 & 12.9 & 13.0 & 8.1 & 11.6 & 0.15M \\
TDF-CNN~\cite{sam2018top} & \textbf{2.7} & 23.4 & 10.7 & 17.6 & \textbf{3.3} & 11.5 &0.13M  \\
C-CNN(ours) & 3.8 & 20.5 & \textbf{8.8} & \textbf{8.8} & 7.7 & \textbf{9.9} & \textbf{0.07M} \\
\hline
CSRNet~\cite{li2018csrnet}  &  2.9  &  \textbf{11.5}  &  \textbf{8.6}  &16.6  & 3.4  &  8.6  &  16.26M \\
SaCNN~\cite{zhang2018crowd}  &  \textbf{2.6} &  13.5 &  10.6  &  12.5 & \textbf{3.3} & \textbf{8.5}  &  24.06M \\
CP-CNN~\cite{sindagi2017generating} &  2.9 &  14.7&  10.5  &  10.4 & 5.8 & 8.86 & 68.40M \\
\hline
\end{tabular}
}
\end{table}

\subsection{Ablation Study}
In this part, we perform an ablation study to analyze the proposed framework on ShanghaiTech Part A dataset.

\begin{figure}[t]
  \centering
  \includegraphics[height=1.5in,width=\linewidth]{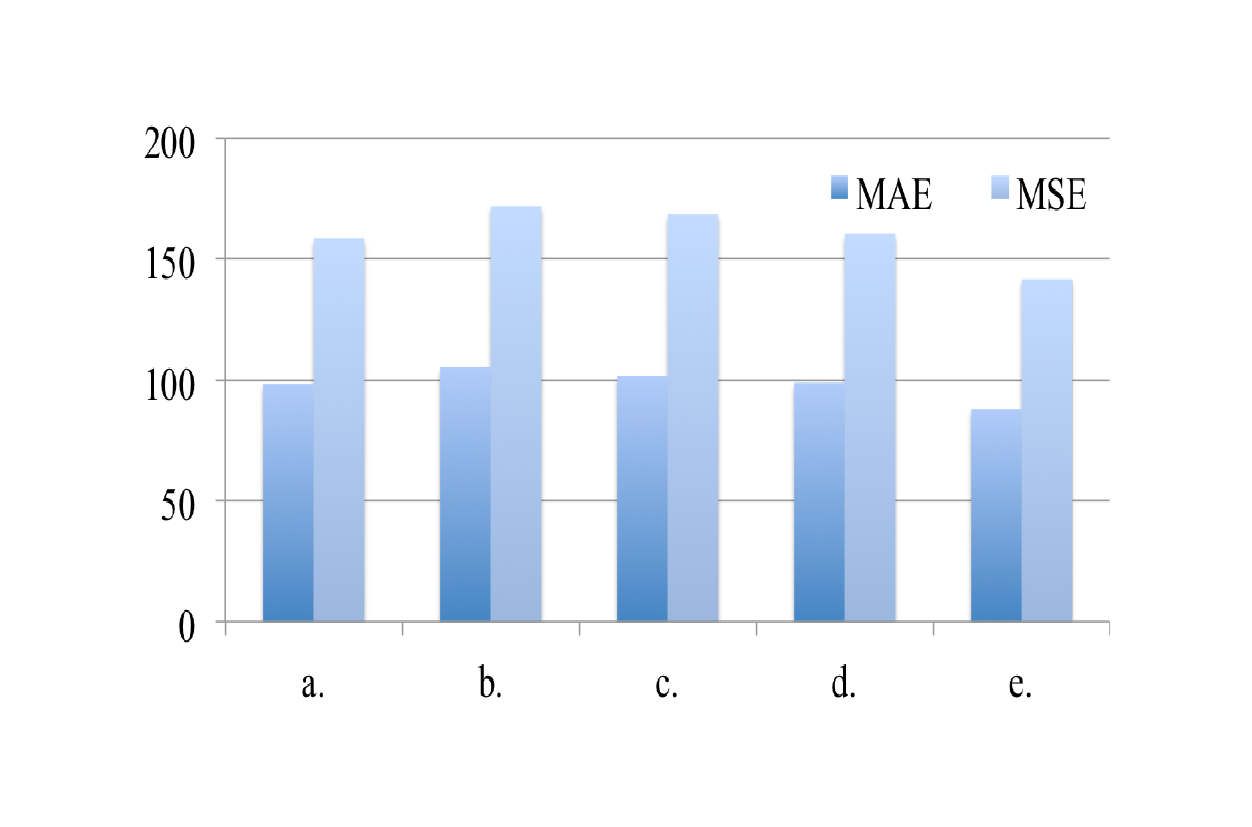}
  \vspace{-0.2in}
  \caption{the effect of varying network architecture, a. 5$\times$5 filter only; b. 7$\times$7 filter only; c. 9$\times$9 filter only;  d. learned weight without last pooling layer; e. proposed approach.}
  \label{fig:ablation}
\end{figure}

\vspace{0.08in}
\noindent\textbf{Network architecture.}
To evaluate the effect of structural variations of the three filters of different sizes, we separately train C-CNN with choice of filters with the size of 5 $\times$ 5, 7 $\times$ 7, 9 $\times$ 9 respectively and all three filters contained concurrently. Figure \ref{fig:ablation} shows the comparison result. We observe that C-CNN with the filter of size 5 $\times$ 5 only outperforms with the filter of 7 $\times$ 7 or the filter of 9 $\times$ 9. It is easy to ascribe to filter of size 5 $\times$ 5 capture crowds at lower scales within the scene and be more advantageous to extract the characteristics of highly congested crowded scenes. With the formation of all three filters contained, the result is 88.08 of MAE and 141.72 of MSE, which is the lowest. It shows that the scale and perspective variations could be adapted better with the structure of three independent filters of respective sizes.

\vspace{0.08in}
\noindent\textbf{Effect of the pooling operation.}
From Figure \ref{fig:ablation} we can observe that without the last pooling operation, we obtain the MAE of 98.9 and MSE of 160.4, which is inferior to the complete model. This evidence tells the fact that the last pooling layer plays an indispensable role in the whole model, and the current architecture has reached its peak of balance between operational speed and prediction accuracy. This is because the pooling layer provides the characteristic of scale invariance to local translation which can help us pay more attention to the existence of the feature other than the exact location of it, especially in the field of crowd counting.

\subsection{Speed Comparison}
In this section, we compare C-CNN with the other two crowd counting methods MCNN and CMTL. The main reason of using MCNN in ~\cite{zhang2016single} and CMTL in ~\cite{sindagi2017cnn} is the relatively small number of parameters provided by the whole network with the lower MAE and MSE displayed in Table \ref{table:shanghai}. Here we use FPS(Frame per second), which is the most commonly used evaluation metrics for measuring the speed of models to access our model fairly. All three methods are tested on ShanghaiTech PartA and PartB, and the figures we reported are calculated running through 182 test images on the ShanghaiTech Part A dataset and 316 test images on the ShanghaiTech Part B dataset at their original resolution (768$\times$1024). Furthermore, these experiments were all conducted under the same condition with a server using the GPU (GeForce GTX 1080) and the CPU (Intel i5-8500 @ 3.00 GHz $\times$ 6). The overall speed comparison with the other state-of-the-art models is demonstrated in Table \ref{table:speed}.

As can be seen from Table \ref{table:speed},
our method gain the highest score at an average speed of 104.16 fps, which is much higher than other methods. with the cost time of 2.14s on the entire testing set from ShanghaiTech Part A and 4.39s on Part B, the speed on prediction phase even achieve more than 10 times faster than that of other advanced models. All these confirm that our work is valuable because only with high speed can real-time processing be realized, which is extremely important for some application scenarios.

\begin{table}
\centering
\caption{The speed of different methods.}
\label{table:speed}
\vspace{-0.1in}
\small{
\begin{tabular}{lccc}
\hline
{Method} &CMTL~\cite{sindagi2017cnn} & MCNN~\cite{zhang2016single}&C-CNN \\  \hline
{FPS}& 8.37&64.52& \textbf{104.16}\\ \hline
\end{tabular}
}
\end{table}

\section{Conclusions}
\label{sec:conclusion}
In this paper, we present a compact CNN for crowd counting to deal with the lack of real-time performance of existing methods. By removing the redundant and recurrent convolutional layers and designing a superior local sparse structure, the parameter size is significantly reduced. Specifically, we using a multiple juxtaposed convolution structure where feature maps extracted from three parallel convolutional layers with different size of receptive fields are directly fused. Compared with the baseline approaches, the proposed model obtains an improvement significantly.

\bibliographystyle{IEEEbib}
\bibliography{refs}

\end{document}